# DeepMatch: Balancing Deep Covariate Representations for Causal Inference Using Adversarial Training

Nathan Kallus

School of Operations Research and Information
Engineering and Cornell Tech,
Cornell University
New York, New York 10011
kallus@cornell.edu

**Abstract**

We study optimal covariate balance for causal inferences from observational data when rich covariates and complex relationships necessitate flexible modeling with neural networks. Standard approaches such as propensity weighting and matching/balancing fail in such settings due to miscalibrated propensity nets and inappropriate covariate representations, respectively. We propose a new method based on adversarial training of a weighting and a discriminator network that effectively addresses this methodological gap. This is demonstrated through new theoretical characterizations of the method as well as empirical results using both fully connected architectures to learn complex relationships and convolutional architectures to handle image confounders, showing how this new method can enable strong causal analyses in these challenging settings.

## 1 Introduction

Drawing causal inferences from observational data often relies on a careful accounting of relevant and systematic differences between treatment and control groups, or else any observed variation in response can be dismissed as spurious correlation rather than a bona-fide causal relationship. For example, hypothetically, differences in lung cancer incidence in coffee drinkers and non-drinkers may be explained away by differing rates of cigarette smoking. To eliminate that possibility, we must contrast groups that are comparable in their smoking rates, *i.e.*, control for that variable. The approach of controlling for confounders relies of course on the assumption that no other unobserved confounders exist, or, more generally and in terms of a causal diagram, that these covariates satisfy the back door criterion (Pearl, 2009).

But controlling for confounders also requires that we understand the way in which they affect treatment, outcome, or both. The rapid development in training neural networks holds promise in enabling us to control for potentially complex relationships and rich covariates with, say, raw image data as in X-rays that are recorded in electronic health records and may go into a doctor's treatment decision and must therefore be accounted for. Standard ways of finding weights that balance covariates for causal inferences rely either on estimating propensity scores (Rosenbaum & Rubin, 1983) or on directly minimizing imbalance metrics, such as kernel maximum mean discrepancy (MMD, Gretton et al., 2009; Kallus, 2016, 2017). But both of these approaches break down when dealing with rich covariates and complex relationships that necessitate flexible modeling with neural networks. Propensity scores estimated by deep neural networks tend to be highly volatile and often miscalibrated as probability estimates. This issue similarly plagues doubly robust approaches (Kang & Schafer, 2007; Robins et al., 1994). At the same time, optimal balancing weights rely very crucially on already having an appropriate representation of the data to balance.

In this paper, we develop a new approach to the problem of balancing covariates in such situations. The approach, termed DeepMatch, is based on solving a game between a weighting and a discriminator network using adversarial training. Underlying this is a new discriminative discrepancy metric that we theoretically characterize and relate to existing metrics used for causal inference. To use it in the context of neural networks requires a few further developments that in the end enable the use of alternating gradient approaches similar to Goodfellow et al. (2016). The method is shown to be statistically consistent. Studying both a case using fully connected networks to learn complex relationships as well as a case using convolutional networks to deal with image confounders, we demonstrate how DeepMatch can enable strong causal analyses in these challenging settings.

## 1.1 Related Literature

There has been intense interest in using machine learning, and neural networks in particular, to estimate causal effects. For the problem of directly regressing individual effects under unconfoundedness, Athey & Imbens (2016); Wager & Athey (2017) study adapting tree-based methods and Johansson et al. (2016); Shalit et al. (2017) study more effective regularization techniques for neural networks. These refine regress-and-compare approaches that ignore covariate shifts. They do not provide covariate balancing weights that can be used for estimating other conditional effects or for doubly robust estimation. Using another identification criterion altogether, Hartford et al. (2017) use neural networks for instrumental-variable analysis, relying on identifying latent natural experiments rather than identifying confounders.

Another strand of work has focused on wrappers that can leverage machine learning predictors as subroutines. Chernozhukov et al. (2016); Hahn (1998) develop doubly robust estimators that use nonparametric estimators such as neural networks to enable efficient estimation of simpler parameters like average effects. Künzel et al. (2017); Nie & Wager (2017) develop meta-learners that combine base learners like neural networks to learn individual causal effects. All of the above rely on having access to weights that balance covariates and generally use estimated propensities. But, as discussed below, deep nets, while good classifiers, yield unwieldy inverse propensity weights. DeepMatch weights may provide a more stable alternative for these wrappers.

A variety of work has recently taken the approach of directly optimizing imbalance metrics for various causal inference tasks (Athey et al., 2016; Bertsimas et al., 2015; Kallus, 2016, 2017, 2018; Zubizarreta, 2012, 2015). For causal estimation from observational data, these take the form of minimizing metrics like Mahalanobis, Wasserstein, and MMD, but, as discussed below, this relies on already having an appropriate representation of the data. DeepMatch proceeds in the same spirit as these, optimizing directly for balance, but uses flexible neural networks to allow for complex relationships and balancing deeper, learned representations.

## 2 Counterfactual Errors, Covariate Balance, and Representations

We consider an observational study consisting of $n$ observations, $\{(X_i, T_i, Y_i^{\text{obs}}) : i = 1, \ldots, n\}$, of the variables $(X, T, Y^{\text{obs}})$, where $X \in \mathcal{X}$ denotes baseline covariates, $T \in \{0, 1\}$ treatment assignment, and $Y^{\text{obs}} \in \mathbb{R}$ observed outcome. For $t = 0, 1$, we let $\mathcal{T}_t = \{i : T_i = t\}$ and $n_t = |\mathcal{T}_t|$. We also let $T_{1:n} = (T_1, \ldots, T_n)$ and $X_{1:n} = (X_1, \ldots, X_n)$ denote all the observed treatment assignments and baseline covariates, respectively. We let $Y_i(0), Y_i(1) \in \mathbb{R}$ be the potential outcomes for unit $i$ so that $Y_i^{\text{obs}} = Y_i(T_i)$. As in Johans-son et al. (2016); Rosenbaum & Rubin (1983) and others, we further assume unconfoundedness, which is key to the identifiability of causal effects:

**Assumption 1.** $\mathbb{E}\left[Y(0) \mid T, X\right] = \mathbb{E}\left[Y(0) \mid X\right]$.

## 2.1 Estimating Causal Effects

The simplest learning task we consider is to estimate the average causal treatment on the treated (ATT):

$$\tau = \mathbb{E}\left[Y(1) - Y(0) \mid T = 1\right] = \mathbb{E}\tau_{n_1},$$
$$\tau_{n_1} = \sum_{i \in \mathcal{T}_1}(Y_i(1) - Y_i(0))$$

The task is nontrivial because only $Y_i(1)$ is observed for treated units and never $Y_i(0)$. This is termed *the fundamental problem of causal inference*: counterfactuals are unobserved. Estimating the ATT essentially relies on a *counterfactual* analysis answering the question "*what would have happened to treated unit i were it administered the control treatment instead?*" Under Asn. 1, this is solved by comparing treated and control units with similar covariates, using regression or matching/weighting or both.

A related task is learning the conditional average causal treatment effect on the treated (CATT):

$$\mathbb{E}\left[Y(1) - Y(0) \mid X^{\mathrm{H}} = x^{\mathrm{H}}, T = 1\right], \qquad (2.1)$$

where $X^{\mathrm{H}}$ is some small subset of interest of the covariates (ATT is a special case with $X^{\mathrm{H}}$ being no covariates). Often, our observational data contains a lot of very rich data $X$ that can help us justify Asn. 1, but only much fewer variables are actually available at run time when we need to make a treatment decision for incoming units, the heterogeneity of the causal effect is only interesting/reasonable along a particular direction of unit difference, and/or we need a simple and interpretable model to understand the heterogeneity explicitly in certain covariates. That is, we *have* to control for all of $X$ to make causal conclusions but we are only interested in effect heterogeneity in $X^{\mathrm{H}}$. We are specifically interested in best-in-class CATT models that, given a class $\mathcal{C}$ (usually of interpretable models like linear), solves

$$\tau_{\mathcal{C}}(\cdot) \in \mathrm{argmin}_{\tau \in \mathcal{C}} \, \mathbb{E}\left[(Y(1) - Y(0) - \tau(X^{\mathrm{H}}))^2 \mid T = 1\right]$$

One can also consider the ATE and CATE – the analogous quantities on the general population without conditioning on $T = 1$ – and all of the methods we discuss easily extend. But, for clarity and brevity, we will focus on ATT and CATT.

Under Asn. 1, many popular estimators for both ATT and CATT are based on the *propensity score* (Rosenbaum & Rubin, 1983). The propensity score function is $e(x) = \mathbb{P}(T = 1 \mid X = x)$ and the propensity score of unit $i$ is $e(X_i)$. In an observational study, the propensity score is generally unknown but it can be estimated by fitting $\hat{e}(x)$ as a probabilistic classifier trained to predict $T$ from $X$, such as logistic regression or a neural network. Given such a $\hat{e}(x)$, a popular estimator for ATT is the inverse probability weighting estimator (IPW) (Horvitz & Thompson, 1952):

$$\hat{\tau}^{\text{IPW}} = \tfrac{1}{n} \sum_{i=1}^{n} (-1)^{1+T_i} W_i^{\text{IPW}} Y_i^{\text{obs}},$$

where $W_i^{\text{IPW}} = T_i + (1 + T_i)\frac{\hat{e}(X_i)}{1-\hat{e}(X_i)}$. If we also have an estimate $\hat{f}_0(x)$ of $f_0(x) = \mathbb{E}[Y(0) \mid X = x]$, then we can form the doubly robust, or augmented IPW (AIPW), estimator Hahn (1998); Robins et al. (1994):

$$\hat{\tau}^{\text{AIPW}} = \tfrac{1}{n} \sum_{i=1}^{n} (-1)^{1+T_i} W_i^{\text{IPW}} (Y_i^{\text{obs}} - \hat{f}_0(X_i)).$$

Similarly, CATT can be estimated using a reweighted loss minimization problem,

$$\hat{\tau}_{\mathcal{C}}^{\text{IPW}}(\cdot) = \operatorname{argmin}_{\tau \in \mathcal{C}} \sum_{i=1}^{n} W_i^{\text{IPW}} (Y_i^{\text{obs}} - \tfrac{2T_i-1}{2}\tau(X_i^{\text{H}}))^2$$

All of these estimators rely on a plug-in approach to dealing with the nuisance parameter $e(x)$ in order to compute an importance weight $W_i^{\text{IPW}}$. Unfortunately, by naïvely plugging in the estimate $\hat{e}(x)$ into the denominator, all these estimators are very susceptible to errors in $\hat{e}(x)$: whenever $1 - e(x)$ is small, even minute errors in $\hat{e}(x)$ can translate to outsize errors in effect estimates. This leads to very volatile estimates with high variance and hence large errors. Practical use requires heuristic stopgaps like trimming and normalizing the weights. The problem is not alleviated either by double robustness alone, as observed by Kang & Schafer (2007) who note that estimators using inverse propensity weights, whether doubly robust or not, are unreliable.

However, this issue is greatly exacerbated when $X$ is very rich so that reliably predicting $T$ from it, and hence estimating $e(x)$, requires a deep neural network. Such models are notorious for overconfidence and miscalibration (Guo et al., 2017). That is, while $\text{sign}(\hat{e}(x) - 1/2)$ may be a good estimate of $\text{sign}(e(x) - 1/2)$, the value of $\hat{e}(x)$ can be far off from $e(x)$ and will generally be either close to 0 or 1, leading to unreliable effect estimates if used as propensities.

In many situations with simple, quantitative covariates, in order to avoid the instability of estimated propensity weights, the standard practice is to use matching Iacus et al. (2012); Rosenbaum (1989), or, more recently, more generalized forms of matching and weighting that leverage modern optimization tools to directly optimize covariate balance Athey et al. (2016); Kallus (2016, 2017); Zubizarreta (2012, 2015).

## 2.2 Optimal Weighting for Covariate Balance

Since estimated propensity weights $W_i^{\text{IPW}}$ can be very volatile as discussed above, we can consider replacing them by other data-driven weights $W$ and analogous estimators:

$$\hat{\tau}_W = \tfrac{1}{n_1} \sum_{i=1}^n (-1)^{1+T_i} W_i Y_i^{\text{obs}} \qquad (2.2)$$

$$\hat{\tau}_{W,\hat{f}_0} = \tfrac{1}{n_1} \sum_{i=1}^n (-1)^{1+T_i} W_i (Y_i^{\text{obs}} - \hat{f}_0(X_i)) \qquad (2.3)$$

$$\hat{\tau}_{W,\mathcal{C}}(\cdot) = \underset{\tau \in \mathcal{C}}{\operatorname{argmin}} \sum_{i=1}^n W_i (Y_i^{\text{obs}} - \tfrac{2T_i-1}{2}\tau(X_i^{\text{H}}))^2 \qquad (2.4)$$

We restrict our attention to weights that correspond to a reweighted control sample, preserving the unit of analysis:

$$W \in \mathcal{W} = \left\{ W \in \mathbb{R}_+^{n_0} : \sum_{i \in \mathcal{T}_0} W_i = n_1 \right\}.$$

Like the estimated propensity weights, we consider choosing the weights $W$ based only on the covariate and treatment data, $W = W(X_{1:n}, T_{1:n})$. Letting $\sigma_0^2(x) = \text{Var}(Y(0) \mid X = x)$, a bias-variance decomposition shows that, under Asn. 1, the risk of $\hat{\tau}_W$ for a particular choice of weights is

$$\mathbb{E}\left[(\hat{\tau}_W - \tau_{n_1})^2 \mid X_{1:n}, T_{1:n}\right] \qquad (2.5)$$
$$= \underbrace{\tfrac{1}{n_1^2}(\sum_{i=1}^n (-1)^{1+T_i} W_i f_0(X_i))^2}_{B^2(W; f_0)} + \underbrace{\tfrac{1}{n_1^2} \sum_{i=1}^n W_i^2 \sigma_0^2(X_i)}_{V^2(W; \sigma_0^2)}.$$

Similar expansions are possible for $\hat{\tau}_{W,\hat{f}_0}$ and $\hat{\tau}_{W,\mathcal{C}}$ (Kallus, 2016). Eq. (2.5) shows that the error in estimating the causal effect is directly related to (a) the balance in covariates as measured by the difference in weighted $f_0$ moments and (b) the $\sigma_0^2$-weighted 2-norm of the weights. However, neither of these are known, suggesting a minimax approach that, given a functional family $\mathcal{F} \subset [\mathcal{X} \to \mathbb{R}]$ and a variance exchange rate $\lambda$ chooses weights $W$ (for any variance bound $\overline{\sigma}^2$) as

$$\underset{W \in \mathcal{W}}{\operatorname{argmin}} \underset{(\lambda f/\overline{\sigma}) \in \mathcal{F}, \sigma^2 \leq \overline{\sigma}^2}{\sup} (B^2(W; f) + V^2(W; \sigma^2))$$

$$= \operatorname{argmin}_{W \in \mathcal{W}} \sup_{f \in \mathcal{F}} B^2(W; f) + \tfrac{\lambda}{n_1^2}\|W\|_2^2. \qquad (2.6)$$

Eq. (2.6) corresponds to minimizing a combination of *balance* and *variance*. This particular minimax approach suggests a balance criterion given by an integral probability metric (IPM, Müller, 1997) distance between the treated and weighted control samples. Given two weighted sets,

$$S^+ = \{(w_1^+, x_1^+), \ldots, (w_{n_+}^+, x_{n_+}^+)\}$$
$$\text{and} \quad S^- = \{(w_1^-, x_1^-), \ldots, (w_{n_-}^-, x_{n_-}^-)\},$$

define the IPM between them with respect to $\mathcal{F}$ as

$$\operatorname{IPM}_{\mathcal{F}}(S^+, S^-) = \sup_{f \in \mathcal{F}} |\textstyle\sum_\pm \sum_{i=1}^{n_\pm} \pm w_i^\pm f(x_i^\pm)|$$

With this definition, eq. (2.6) is equivalent to

$$\min_{W \in \mathcal{W}} \operatorname{IPM}_{\mathcal{F}}^2(\{(\tfrac{1}{n_1}, X_i)\}_{i \in \mathcal{T}_1}, \{(\tfrac{W_i}{n_1}, X_i)\}_{i \in \mathcal{T}_0}) + \tfrac{\lambda}{n_1^2}\|W\|_2^2$$

## 2.3 The Role of a Representation

Popular examples of IPMs include the total variation (TV) distance, given by the IPM for all functions point-wise bounded by 1; the MMD (Gretton et al., 2006), given by the IPM for the unit ball of a reproducing kernel Hilbert space (RKHS); and the Wasserstein distance, given by the IPM for all 1-Lipschitz functions. The MMD and Wasserstein distances have been popular for handling both matching for causal inference and for covariate shift (Gretton et al., 2009; Kallus, 2016). Specifically, Kallus (2016) proposes eq. (2.6) as a weighting objective for estimators of the form eqs. (2.2)–(2.3) and shows that solving eq. (2.6) with the Wasserstein metric and $\lambda \to 0$ corresponds exactly to the classic pairwise matching approach to causal estimation (Rosenbaum, 1989). The TV distance, on the other hand, is uninformative for measuring covariate balance for non-discrete data: since it corresponds to almost-everywhere point-wise differences between measures, it is always equal 2 regardless of $w_i^s$ whenever $\{x_i^+\} \neq \{x_i^-\}$ and is therefore not useful for finding covariate-balancing weights.

But balancing covariates based on MMD and Wasserstein distances delicately relies on having an appropriate representation of $X$ for the task. For example, the kernel MMD with a kernel $K$ would represent $X$ as $K(X, \cdot)$ and measure distances as the 2-norm distance between the weighted sample means in this representation. But this representation may be inappropriate for very rich covariates, *e.g.* containing image data. Similarly, Wasserstein distances lead to weights given by pairwise matching, which would seek matches based on Euclidean distances in image pixels. Both of these are wholly inappropriate for such rich data because they rely on a functional family $\mathcal{F}$ without the right structure – loosely including all Lipschitz functions or misguidedly relying on smoothness in raw pixels, or other very rich data structure.

One possible solution may be to first put the covariate data into an appropriate representation. But this requires learning such a representation and will rely on its sufficiency for the causal estimation task. For example, for covariate data given by images, one would first have to use auxiliary labeled data to learn a convolutional network and extract from it an appropriate hidden representation. Even if such auxiliary data is available, there is no guarantee that this representation is also an appropriate one for the causal estimation task. In particular, after this preprocessing we may lose unconfoundedness due to omitting important aspects of the data and due to estimation errors in learning the representation.

Instead, in this paper, we will *directly* imbue $\mathcal{F}$ with the structure of a (potentially deep) neural network, such as a convolutional neural network (CNN). This will induce a less loose structure more appropriate for the task and lead to the simultaneous learning and balancing of a sufficient representation of the covariate data.

However, optimizing eq. (2.6) with respect to such functional families is problematic – the corresponding IPM is not closed form and using alternating saddle-point-finding methods directly on eq. (2.6) fail. We will instead develop a new, discriminative model for covariate balance, relate it the one based on IPMs, and use adversarial training in order to optimize it. This will lead to an alternative way to find weights that more resembles the generator-discriminator game of Goodfellow et al. (2014).

## 3   A Discriminative Model of Covariate Balance

For $f : \mathcal{X} \to \mathbb{R}$, consider the binary classifier that predicts the probability of a positive given $x$ as $\text{logit}^{-1}(f(x))$ where $\text{logit}^{-1}(v) = 1/(1+\exp(-v))$. Letting $\ell(z) = \log(\text{logit}^{-1}(z)) + \log 2$, and given weighted positive and negative samples, $S^+$ and $S^-$, the log likelihood of the classifier relative to the random classifier is then

$$L(f; S^+, S^-) = \sum_{\pm} \sum_{i=1}^{n^{\pm}} w_i^{\pm} \ell(\pm f(x_i^{\pm})).$$

$L(f; S^+, S^-)$ with $w_i^{\pm} = 1$ is the training objective for both logistic regression and neural network classifiers. Similarly, training generative adversarial net-works corresponds to having $w_i^\pm = 1$ and letting $x_i^+$ be given data and $x_i^-$ be generated by a generator network with random inputs. Instead, we are here interested in training the weights $w_i^\pm$, leaving *both* $x_i^+$ and $x_i^-$ as fixed, given data. Maximizing $L(f; S^+, S^-)$ over *all* $f(x)$ yields the Jensen-Shannon (JS) divergence, as observed by Goodfellow et al. (2014). But, like the TV distance, the JS divergence is uninformative for balancing when $\{x_i^+\} \neq \{x_i^-\}$ (always 1, regardless of $w_i^\pm$). Instead, we will be interested in restricting $\mathcal{F}$ in a meaningful way in order to come up with a balancing metric more similar to IPM but based on a discriminative objective.

Toward that end, define more generally the squared $\psi$-*discriminative distance* (for $\psi \geq 0$) with respect to $\mathcal{F}$ between the weighted samples $S^+, S^-$ as

$$\mathrm{DD}^2_{\mathcal{F}, \psi}(S^+, S^-) = \sup_{f \in \mathcal{F}, t \in \mathbb{R}} L_\psi(f, t; S^+, S^-), \qquad (3.1)$$
$$L_\psi(f, t; S^+, S^-) = L(tf; S^+, S^-) - \tfrac{\psi}{2} t^2$$

In particular, if $\mathcal{F}$ are all neural networks of a given architecture with sum squared weights no more than 1, then eq. (3.1) corresponds to training a neural-net binary classifier with $\psi$ weight decay. If $\mathcal{F}$ is the unit ball of a RKHS then eq. (3.1) corresponds to kernelized logistic regression (Jaakkola & Haussler, 1999). Standard logistic regression (with possible ridge regularization $\psi$) is a special case of either of these (a one-layer net or a linear kernel logistic regression).

The discriminative distance, defined as the square root of eq. (3.1) is well-defined:

**Theorem 1.** $\mathrm{DD}^2_{\mathcal{F}, \psi}(S^+, S^-)$ *is finite nonnegative. Consequently, it has a well-defined square root,* $\mathrm{DD}_{\mathcal{F}, \psi}(S^+, S^-)$.

Unlike the IPM, the discriminative distance is *not* a pseudo-metric because it does not satisfy the triangle inequality:

**Example 1.** Let $S^{(1)} = \{(1, 1)\}$, $S^{(2)} = \{(1, -1)\}$, $S^{(3)} = \{(1/2, 1), (1/2, -1)\}$, and $\mathcal{F} = \{x \mapsto \pm x\}$. Then $\mathrm{DD}_{\mathcal{F}, 0}(S^{(1)}, S^{(2)}) = \sqrt{2 \log(2)} > \sqrt{2 \log(27/16)} = \mathrm{DD}_{\mathcal{F}, 0}(S^{(1)}, S^{(3)}) + \mathrm{DD}_{\mathcal{F}, 0}(S^{(3)}, S^{(2)})$. In fact, the inequality is strict for all $\psi \geq 0$.

## 3.1 Characterizing the Discriminative Distance

Next, we argue that the discriminative distance, while not a metric, is related to, and sometimes similar to, the IPM. We first give a dual characterization of the discriminative distance that relates it directly to the IPM over the same $\mathcal{F}$.

**Theorem 2.** *Let $h(p) = p \log(p) + (1-p)\log(1-p) + \log 2$. Suppose $\mathcal{F}$ is a symmetric convex set. (i.e., for all $f, f' \in \mathcal{F}$, $p \in [0,1]$, and sign $\pm$, we have $pf \pm (1-p)f' \in \mathcal{F}$).*

*If $\psi > 0$ then $\mathrm{DD}^2_{\mathcal{F},\psi}(S^+, S^-)$ is equal to*

$$\inf_{0 \leq \mathbf{p} \leq 1} \left( \sum_{i=1}^{n_+} w_i^+ h(p_i^+) + \sum_{i=1}^{n_-} w_i^- h(p_i^-) \right. \quad (3.2)$$
$$\left. + \tfrac{\psi^{-1}}{2} \mathrm{IPM}^2_{\mathcal{F}}(\{(p_i^+ w_i^+, x_i^+)\}_{i=1}^{n_+}, \{(p_i^- w_i^-, x_i^-)\}_{i=1}^{n_-}) \right)$$

*And, if $\psi = 0$ then $\mathrm{DD}^2_{\mathcal{F},\psi}(S^+, S^-)$ is equal to*

$$\inf_{0 \leq \mathbf{p} \leq 1} \sum_{i=1}^{n_+} w_i^+ h(p_i^+) + \sum_{i=1}^{n_-} w_i^- h(p_i^-) \quad (3.3)$$
$$\text{s.t. } \mathrm{IPM}_{\mathcal{F}}(\{(p_i^+ w_i^+, x_i^+)\}_{i=1}^{n_+}, \{(p_i^- w_i^-, x_i^-)\}_{i=1}^{n_-}) = 0$$

This result can be used to show that the discriminative distance is zero precisely when the corresponding IPM is zero.

**Theorem 3.** $\mathrm{DD}_{\mathcal{F},\psi}(S^+, S^-) = 0 \iff \mathrm{IPM}_{\mathcal{F}}(S^+, S^-) = 0.$

However, when $\psi = 0$, they need not induce the same topology. The next example shows that the IPM can approach zero while the discriminative distance remains constant:

**Example 2.** Fix $\delta > 0$. Let $S^+ = \{(1, \delta/2)\}$, $S^- = \{(1, -\delta/2)\}$, and $\mathcal{F} = \{x \mapsto \pm x\}$. Then we have that $\mathrm{IPM}_{\mathcal{F}}(S^+, S^-) = \delta$ is arbitrarily small while $\mathrm{DD}_{\mathcal{F},0}(S^+, S^-) = \sqrt{2\log(2)}$ is bounded away from zero.

But when $\psi > 0$, we can show that they *will* in fact induce the *same topology* on any space of weighted sets such that

$$M = \sup_{\pm, i \leq n^\pm, f \in \mathcal{F}} |f(x_i^\pm)| \quad \text{and} \quad \overline{w} = \sum_\pm \sum_{i=1}^{n^\pm} w_i^\pm$$

are bounded. Usually, we will have $\overline{w} = 2$ since each set of weights sums to 1. If $\mathcal{F}$ is all all neural networks of a given architecture (including a one-layer linear function) with sum squared weights no more than 1, then $M$ is bounded over points $x$ of bounded norm. $M$ is also bounded if $\mathcal{F}$ is the unit ball of an RKHS with a bounded kernel (*e.g.*, RBF). Specifically, we can bound the ratio between the two distances (recall that either both or neither are zero):

**Theorem 4.** $2\sqrt{2\psi} \leq \frac{\mathrm{IPM}_{\mathcal{F}}(S^+, S^-)}{\mathrm{DD}_{\mathcal{F},\psi}(S^+, S^-)} \leq \max(2M\sqrt{\overline{w}}, 4\sqrt{\psi}).$

Moreover, we can show that, as $\psi$ grows, the IPM and the discriminative distance become the same up to scaling.

**Theorem 5.** *For any weighted sets $S^+, S^-$,*

$$\lim_{\psi \to \infty} 2\sqrt{2\psi}\, \mathrm{DD}_{\mathcal{F},\psi}(S^+, S^-) = \mathrm{IPM}_{\mathcal{F}}(S^+, S^-).$$

*The limit holds uniformly over $S^+, S^-$ with bounded $\overline{w}, M$.*

**Algorithm 1** Conditional Gradient for Eq. (3.4)
>   **input**: $X_{1:n}$, $T_{1:n}$, $\lambda$, $K$, and an oracle as in eq. (3.5)
>   Set $W_i = 1/n_0$ for all $i \in \mathcal{T}_0$
>   **for** $k = 1, \ldots, K$ **do**
>     Set $S^+ = \{(\frac{1}{n_1}, X_i)\}_{i \in \mathcal{T}_1}$ and $S^- = \{(\frac{W_i}{n_1}, X_i)\}_{i \in \mathcal{T}_0}$
>     Get the corresponding $f^{\text{oracle}}, t^{\text{oracle}}$ in eq. (3.5)
>     Let $i = \text{argmin}_{i \in \mathcal{T}_0}(\ell(-t^{\text{oracle}} f^{\text{oracle}}(X_i)) + \frac{2\lambda}{n_1^2} W_i)$
>     $W \leftarrow (k-1)/(1+k)W$, $W_i \leftarrow W_i + 2/(1+k)$
>   **end for**
>   **output**: $W$

## 3.2 Optimizing the Discriminative Distance

The last section suggests that the discriminative distance with respect to $\mathcal{F}$ can be used as a surrogate for covariate balance. In particular, measuring covariate imbalances using the discriminative distance, we have shown that eliminating the discriminative distance when $\mathcal{F}$ consists of neural networks of a given architecture corresponds exactly to eliminating *any* estimation bias due to imbalances for any outcomes that can be well-approximated by such neural networks. Therefore, an alternative criterion for choosing weights is to minimize imbalance as measured by the discriminative distance plus a potential variance regularizer:

$$W^* \in \text{argmin}_{W \in \mathcal{W}}(I^2(W) + \frac{\lambda}{n_1^2}\|W\|_2^2) \quad (3.4)$$
$$I^2(W) = \text{DD}^2_{\mathcal{F},\psi}(\{(\tfrac{1}{n_1}, X_i)\}_{i \in \mathcal{T}_1}, \{(\tfrac{W_i}{n_1}, X_i)\}_{i \in \mathcal{T}_0})$$

The question then is how to find such weights $W$ that optimize eq. (3.4). We first show the problem is convex.

**Theorem 6.** *The optimization problem eq. (3.4) is convex for any $\mathcal{F}$, $\psi$, and $\lambda \geq 0$.*

To optimize eq. (3.4), we first consider the simple case where we have an oracle for the optimization problem eq. (3.1). Specifically, suppose that, given $S^+, S^-$ we could easily find $f^{\text{oracle}} \in \text{cl}(\mathcal{F})$, $t^{\text{oracle}} \in \mathbb{R}$ such that

$$L_\psi(f^{\text{oracle}}, t^{\text{oracle}}; S^+, S^-) = \text{DD}^2_{\mathcal{F},\psi}(S^+, S^-). \quad (3.5)$$

For example, in the case of $\mathcal{F}$ being linear functions or being an RKHS, this amounts to solving logistic regression (possibly kernelized), which can be done quickly and easily for even large (but not huge) problems (Jaakkola & Haussler, 1999). Given such an oracle, we can employ the conditional gradient algorithm (Frank & Wolfe, 1956; Jaggi, 2013) to solve eq. (3.4), producing Algo. 1. Note that we can employ Thm. 2 to show that eq. (3.4) can also be written as a single convex minimization problem in $W$ and $\mathbf{p}$ with an IPM in the objective (or, in a constraint for $\psi = 0$). Therefore, we can also solve eq. (3.4) using an oracle for the IPM instead, albeit with slightly more complicated gradients and twice the number of optimization variables.

### 3.3 A Shallow Example

Before proceeding to develop Deep-Match, we first consider a shallow example, that is, one with few dimensions and simple treatment and outcome models, in order to illustrate the discriminative distance and the weight-optimizing Alg. 1. We consider 2-dimensional covariates drawn uniformly on the unit square, $X_{ij} \sim \text{Uniform}[-1, 1]$ for $i = 1, \ldots, n$, $j = 1, 2$. We let treatment be more likely above the diagonal $x_1 = -x_2$ line and less likely below it: $e(X_i) = 0.1$ if $X_{i1} + X_{i2} < 0$ and otherwise $e(X_i) = 0.9$. And, we let outcomes be exponential in covariates and assume no effect: $Y_i(0) = Y_i(1) = e^{X_{i1}+X_{i2}} + \epsilon_i$, where $\epsilon_i \sim \mathcal{N}(0, 1)$.

Fixing a particular draw of $n = 300$ units, we plot the covariates in Fig. 1(a). Letting $\mathcal{F} = \{x \mapsto \alpha + \beta^T x : \|\beta\|_2 \leq 1\}$, we compute the corresponding IPM (which amounts to Euclidean distance between means) and discriminative distance (which is given by an $\ell_2$-regularized logistic regression) between the raw (not reweighted) con-

Figure 1: Results in the illustrative shallow example

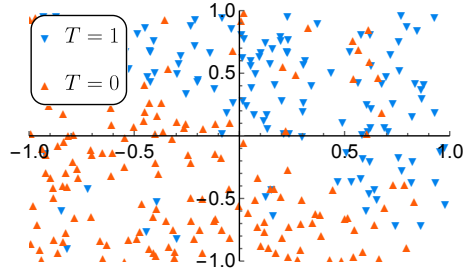

(a) 2-dimensional covariates $X_{1:n}$

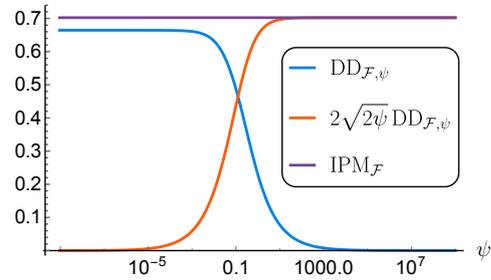

(b) The discriminative distance and IPM

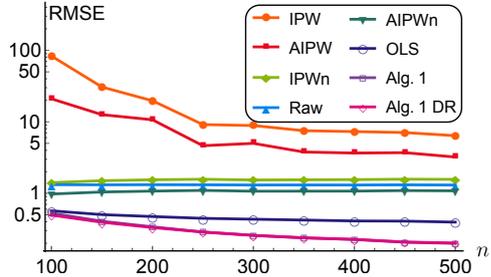

(c) RMSE of different ATT estimators

trol and treated samples. Varying $\psi \geq 0$, we plot the results in Fig. 1(b) along with the $2\sqrt{2\psi}$-scaled discriminative distance as in Thms. 4 and 5. We see that, as the theorems promise, the scaled discriminative distance converges to the IPM from below.

Next, we consider how well Alg. 1 balances covariates for estimating causal effects. For each of $n = 100, 150, \ldots, 500$, we compute the RMSE of 8 different estimators over 2000 simulations and plot the results in Fig. 1(c). The IPW estimator is given by propensities estimated by logistic regression and IPWn is given by normalizing the IPW weights to sum to 1, which we see provides some improvement. AIPW and AIPWn are the corresponding doubly robust estimators with $\hat{f}_0$ estimated by OLS, which respectively improves on each of unaugmented estimators. We also show the errors of the raw (unweighted) mean difference, which beats the IPW, IPWn, and AIPW estimates. An OLS estimate, which is also equivalent to using unconstrained weights that minimize the IPM (Kallus, 2016, Thm. 13), beats all propensity-based estimates. Finally, we compute optimal weights to minimize the discriminative distance ($\psi = 0$) using Alg. 1 ($\lambda = 1$) and consider using both the vanilla weighted estimator $\hat{\tau}_W$ and the doubly robust estimator $\hat{\tau}_{W,\hat{f}_0}$. This yields the least error, with double robustness giving marginal improvements for small $n$.

## 4 Going Deep

We are now prepared to develop the DeepMatch method, which seeks balance complex and deep representation of the covariate data. DeepMatch does this by seeking weights that minimize the discriminative distance with respect to a deep discriminator, *i.e.*, when $\mathcal{F}$ is given by all neural networks of a given architecture. There are two barriers to doing this with the tools we developed so far (Alg. 1). First, as $n$ grows, additional runs of the conditional gradient algorithm are necessary to generate a dense set of weights, since the number of nonzero weights is bounded above by the number of iterations. Second, each of these runs becomes increasingly difficult to compute when $\mathcal{F}$ is complex, as in the case of a deep neural network. Computing the corresponding oracle can be exceedingly difficult because, not only is optimizing a neural network taxing, one usually does not optimize it fully, even to mere local optimality, and so the weight gradient recovered may be highly inaccurate.

Instead, DeepMatch will rely on similar alternating descent methods as are used in adversarial training of GANs (Goodfellow et al., 2014) to solve minimax problems with neural nets. That is, we will address problem eq. (3.4) by taking alternating (or, simultaneous) gradient steps in $W \in \mathcal{W}$ and $f \in \mathcal{F}$, each while treating the other as fixed. However, to apply this approach here, we will need to make further developments to make our problem amenable to the standard solution methods that use stochastic optimization in order to deal with large datasets and complex objectives. In particular, we need to be able to solve the problem in *mini-batches*. To do this, we first show we can relax constraints that link the optimization variables across data points and we then show how we can parametrize the weights using their own neural network.

## 4.1 Relaxing Weight Normalization

One impediment to applying mini-batched descent to problem eq. (3.4) is the normalization constraint, that weights have to sum to a certain fixed value, which links the optimization variables across data points $i$. Not only does this make stochastic mini-batching difficult, the constraint also becomes non-convex if we parametrize the weights as we are going to do in the next section. Therefore, the first step is to find an alternative way to enforce the normalization constraint. The next theorem will guide the approach.

**Theorem 7.** *Problem eq. (3.4) is equivalent to*

$$\min_{\phi \geq 0, W \in \mathcal{W}(\phi)} I^2\left(\frac{n_1 W}{\sum_{i \in \mathcal{T}_0} W_i}\right) + \lambda \frac{\sum_{i \in \mathcal{T}_0} W_i^2}{(\sum_{i \in \mathcal{T}_0} W_i)^2}, \text{ where} \quad (4.1)$$

$$\mathcal{W}(\phi) = \underset{W \geq 0}{\operatorname{argmin}} I^2(W) + \sum_{i \in \mathcal{T}_0}\left(\lambda \frac{W_i^2}{n_1^2} + \phi \frac{W_i}{n_1}\right). \quad (4.2)$$

Thm. 7 suggests the following approach to optimizing eq. (3.4): select a grid of $\phi$ values, $\phi_1, \ldots, \phi_K$; solve eq. (4.2) for each $\phi$ value; normalize the resulting optimal set of weights; and chose the best set of weights by plugging each into the objective of eq. (3.4). We discuss the details of optimizing eq. (4.2) and choosing the $\phi$ grid in Sec. 4.3.

## 4.2 Parameterizing $W$ and Choosing an Activation

Next, we parameterize the weights $W$ using their own neural network. That is, we replace the $n_0$ optimization variables given by $W$ by a neural network that produces their values. A question that arises is what activation to choose for the network's output. Choosing a nonnegative activation, such as a ReLU, easily accommodates the constraint that the weights be nonnegative. However, it may produce weights that are too sparse, outputting many zeros. Other activations such as logistic or $1 + \tanh$ are everywhere positive but have limited range, while the weight variables in eq. (4.2) are unrestricted and unnormalized.

Instead, we take inspiration from the weight function that arises from a direct application of inverse propensity weighting with propensities that would be estimated by logistic regression or a neural network. Estimating the propensity function as $\hat{e}(x) = \text{logit}^{-1}(g(x))$, for any function $g$, leads to the plug-in importance weight $W_i^{\text{IPW}} = \frac{\hat{e}(X_i)}{1-\hat{e}(X_i)} = \exp(g(x))$. As previously discussed and illustrated, such plug in weights lead to unwieldy and inaccurate estimates, but, as a functional form, this suggests exp as the activation on the output layer for any weight-producing network, which can then be trained to produce weights directly optimized for accuracy. Considering some parametric family $g(x; \theta_g)$, such as neural networks of a given architecture, we will solve eq. (4.2) by transforming $W_i = \exp(g(X_i; \theta_g))$ and optimizing over $\theta_g$ instead of over $W \in \mathcal{W}$.

## 4.3 Putting It Together

We now use the insights of the last two sections to develop the DeepMatch method. Let us express $\mathcal{F}$ parametrically and assume it has the form $\mathcal{F} = \{f(\cdot; \theta_f) : R(\theta_f) \leq 1\}$ for some degree-2 homogeneous regularizer $R$ (can be zero). And let us write $\ell_i(\theta_f) = \ell((-1)^{T_i+1} f(X_i; \theta_f))$. Then solving the parametrized version of eq. (4.2), where we set $W_i = \exp(g(X_i; \theta_g))$, amounts to solving the following zero-sum game in $\theta_g$ and $\theta_f$:

$$\min_{\theta_g} \max_{\theta_f} L_\phi(\theta_g, \theta_f), \quad \text{where} \tag{4.3}$$
$$L_\phi(\theta_g, \theta_f) = \tfrac{1}{n_1} \sum_{i \in \mathcal{T}_1} \ell_i(\theta_f) + \tfrac{1}{n_1} \sum_{i \in \mathcal{T}_0} e^{g(X_i;\theta_g)} \ell_i(\theta_f)$$
$$- \tfrac{\psi}{2} R(\theta_f) + \tfrac{\lambda}{n_1^2} \sum_{i \in \mathcal{T}_0} e^{2g(X_i;\theta_g)} + \tfrac{\phi}{n_1} \sum_{i \in \mathcal{T}_0} e^{g(X_i;\theta_g)}$$
$$= \sum_{i=1}^n u_i(\theta_f, \theta_g; \psi, \lambda, \phi) - \tfrac{\psi}{2} R(\theta_f),$$

where each $u_i$ depends on the data $X_i, T_i$. Note that if $R(\theta_f)$ is convex then $L(\theta_g, \theta_f)$ is convex in $\theta_g$ and concave in $\theta_f$. Consequently, by the von Neumann minimax theorem, the min and max can be exchanged (under some regularity). Note also that $L(\theta_g, \theta_f)$ is of the form of a sum of loss functions separable over the data plus some regularization (we can also easily include regularization in $\theta_g$).

Using this formulation, we can develop Alg. 2 to find optimal weights for balancing covariates with respect to the discriminative distance over neural network discriminators. For each value of $\phi$, the algorithm proceeds in two stages. In the first stage we address eq. (4.2) by seeking an equilibrium to eq. (4.3). To do this, over $K_1$ epochs, we cycle through mini-batches of size $B$ and, for each, we let $\theta_f$ ascend its gradient and $\theta_g$ descend its (we

---

**Algorithm 2** DeepMatch

---
**input**: $X_{1:n}$, $T_{1:n}$, $\psi$, $\lambda$, $K_1$, $K_2$, $B$, $M$, $\Phi$, regularizer $R$, network architectures, and a gradient update rule
$v^* \leftarrow \infty$
**for** $\phi \in \Phi$, $m = 1, \ldots, M$ **do**
    Randomly initialize the parameters $\theta_f, \theta_g$ of the discriminator and weight networks
    **for** $k = 1, \ldots, K_1$ **do**
        Shuffle the data into mini-batches $I_1, \ldots, I_{\lceil n/B \rceil}$ of sizes $|I_j| \in \{B, B-1\}$
        **for** $j = 1, \ldots, \lceil n/B \rceil$ **do**
            $\delta_g \leftarrow \sum_{i \in I_j} \nabla_{\theta_g} u_i(\theta_f, \theta_g; \psi, \lambda, \phi)$
            $\delta_f \leftarrow \sum_{i \in I_j} \nabla_{\theta_f} u_i(\theta_f, \theta_g; \psi, \lambda, \phi) - \frac{\psi}{2} \nabla R(\theta_f)$
            Move $\theta_f$ in direction $\delta_f$ and $\theta_g$ in direction $-\delta_g$
        **end for**
    **end for**
    **for** $i = 1, \ldots, n$ **do** $W_i \leftarrow T_i + (1 - T_i) \frac{n_1 e^{g(X_i; \theta_g)}}{\sum_{j \in \mathcal{T}_0} e^{g(X_j; \theta_g)}}$
    **for** $k = 1, \ldots, K_2$ **do**
        Shuffle the data into mini-batches $I_1, \ldots, I_{\lceil n/B \rceil}$
        **for** $j = 1, \ldots, \lceil n/B \rceil$ **do**
            $\delta_f \leftarrow \frac{1}{n_1} \sum_{i \in I_j} W_i \nabla \ell_i(\theta_f) - \frac{\psi}{2} \nabla R(\theta_f)$
            Move $\theta_f$ in direction $\delta_f$
        **end for**
    **end for**
    $v \leftarrow \frac{1}{n_1} \sum_{i \in I_j} W_i \ell_i(\theta_f) - \frac{\psi}{2} R(\theta_f) + \lambda \sum_{i \in \mathcal{T}_0} W_i^2$
    **if** $v^* > v$ **then** $v^* \leftarrow v, W^* \leftarrow W$
**end for**
**output**: $W^*$

---

use simultaneous updates; alternating is another option). We can use any stochastic gradient update rule. In our experiments, we use Adam (Kingma & Ba, 2014) with a global learning rate of $10^{-4}$ (for other options see Goodfellow et al., 2016, Ch. 8). In the second stage, we address eq. (4.1) by trying to evaluate the discriminative distance of the resulting weights after normalization. To do this, we simply fix the weights and train *only* the discriminator in this stage, over $K_2$ epochs with mini-batches of size $B$. Because even for a single $\phi$ the first stage can end up at different weights, we do several ($M$) runs of each $\phi$. Finally, we make sure to keep the weights over all these runs with the best objective so far in eq. (4.1).

A remaining detail is how to choose the grid $\Phi$ of $\phi$ values. Due to convexity, Thm. 7 also gives that the sum of weights of solutions $\mathcal{W}(\phi)$ is monotonic in $\phi$ whereas the optimizer of eq. (3.4) is given by the root where the sum of weights is exactly $n_1$. We directly compare objective rather than use a line search over Lagrange multipliers because our optimization for each $\phi$ is inexact. Nonetheless, we can use this to find an appropriate range for $\phi$. Picking a tolerance $\eta \in (0, 1)$, we use binary search to find values $\underline{\phi}, \overline{\phi}$ that give weights with sum $\geq \eta n_1$ and $\leq n_1/\eta$, respectively. Then, we divide the values in between into a grid $\Phi$.

### 4.4 Theoretical Characterization

We next prove that if we were to do the optimization exactly, then we indeed estimate the causal effects of interest. The rates will depend on the Rademacher complexity of $\mathcal{F}$:

$$\mathfrak{R}_n(\mathcal{F}) = \tfrac{1}{2^n} \sum_{\xi \in \{\pm 1\}^n} \sup_{f \in \mathcal{F}} \left| \tfrac{1}{n} \sum_{i=1}^n f(X_i; \theta_f) \right|.$$

For both linear models and neural networks, $\mathfrak{R}_n(\mathcal{F}) = O_p(1/\sqrt{n})$ (Bartlett et al., 2017; Kakade et al., 2009).

**Theorem 8.** *Let* $\Theta_g = \bigcup_{\phi > 0} \operatorname{argmin}_{\theta_g} \max_{\theta_f} L_\phi(\theta_g, \theta_f)$,

$$W_i(\theta_g) = T_i + (1 - T_i) \frac{n_1 e^{g(X_i; \theta_g)}}{\sum_{j \in \mathcal{T}_0} e^{g(X_j; \theta_g)}},$$
$$I^2(W) = \sup_{\theta_f} \tfrac{1}{n_1} \sum_{i=1}^n W_i \ell_i(\theta_f) - \tfrac{\psi}{2} R(\theta_f), \text{ and}$$
$$W^* \in \operatorname{argmin}_{W \in W(\Theta_g)} I^2(W) + \lambda \sum_{i \in \mathcal{T}_0} W_i^2.$$

*Suppose that Asn. 1 holds, $\psi > 0$, $\lambda > 0$, $\sigma_0^2(X)$ is a.s. bounded, $e(X)$ is a.s. bounded away from 1, $\exists \theta_f : f_0(x) = f(x; \theta_f)$, and $\exists \theta_g : e(x) = \operatorname{logit}^{-1}(g(x; \theta_g))$. Then,*

$$\hat{\tau}_{W^*} - \tau = O_p(\mathfrak{R}_n(\mathcal{F}) + 1/\sqrt{n})$$

## 5 Experiments

### 5.1 A Fully Connected Example

We begin with an example involving fully connected networks. First, we set up the data-generating process. We consider 6-dimensional covariates $X_{ij} \sim \text{Uniform}[-2, 2]$ for $i = 1, \ldots, n$, $j = 1, \ldots, 6$ and a somewhat complex treatment model involving the XOR of their sign $e(X_i) = 0.05$ if $\bigoplus_{j=1}^{6}(X_{ij} > 0)$ and otherwise $e(X_i) = 0.95$ (where $\oplus$ is the XOR). We let outcomes be exponential in covariates: $Y_i(t) = e^{\sum_{j=1}^{6} X_{ij}} + T_i \sum_{j=1}^{6} X_{ij} - T_i + \epsilon_i$ where $\epsilon_i \sim \mathcal{N}(0,1)$. We consider two tasks: estimating ATT and learning the CATT in $X^H = \sum_{j=1}^{6} X_{ij}$. For estimating ATT we consider several strategies: a regression approach based on estimating $\hat{f}_0(x)$ using a neural network, an IPW approach based on estimating $\hat{e}(x)$ using a neural network, an AIPW approach combining these, and the same with normalized propensities (A/IPWn). We compare these to DeepMatch with the same network architectures, $\psi = 0$, and $\lambda$ either 0 or 1. All neural networks are fully connected with 4 hidden layers of 2 neurons each and ReLU activations. Each network has a different and appropriate output activation and loss. We use $K_1 = 10$ epochs with mini-batches of 100 to train all networks and for the first stage of DeepMatch and we use $K_2 = 5$ epochs for the second stage. We use $M = 5$ and a grid of 50 $\phi$ values based on $\eta = 0.01$.

Table 1: Results of the fully connected example

| | (a) ATT | | | (b) CATT | | |
|---|---|---|---|---|---|---|
| Method | Bias | SE | RMSE | Bias | SE | RMSE |
| Raw | -8.33 | 6.44 | 10.53 | -10.22 | 8.35 | 13.20 |
| IPW | -8.25 | 6.29 | 10.38 | -10.19 | 7.77 | 12.82 |
| IPWn | -8.34 | 6.37 | 10.50 | -10.26 | 7.84 | 12.91 |
| Regn | 9.42 | 4.00 | 10.24 | 8.37 | 2.40 | 8.71 |
| AIPW | -8.07 | 6.22 | 10.19 | – | – | – |
| AIPWn | -8.14 | 6.30 | 10.30 | – | – | – |
| $DM_0$ | -0.45 | 4.26 | 4.28 | -2.70 | 4.54 | 5.29 |
| $DM_{\frac{1}{2}}$ | 0.85 | 3.89 | 3.98 | -1.25 | 3.68 | **3.89** |
| $DM_0$ DR | -0.55 | 4.03 | 4.06 | – | – | – |
| $DM_{\frac{1}{2}}$ DR | 0.62 | 3.76 | **3.82** | – | – | – |

We run 50 replications of the experiment with $n = 1000$ and report the bias, std. error, and RMSE of each in Tab. 1(a). We see that inverse-propensity-based approaches yield very volatile estimates, where normalizing or augmenting does help but only marginally. Using DeepMatch (DM) reduces MSE significantly. Combining it with the neural network regression in a doubly robust estimator (DM-DR) reduces error slightly further. Using $\lambda > 0$ provides somewhat better results by controlling variance, but does not make a substantial difference in MSE. Next, we estimate the CATT in $X^H$ as a linear function. We consider solving eq. (2.4) with uniform weights, IPW weights, or DeepMatch weights or minimizing an empirical average of eq. (2.1) with regression-imputed outcomes. We report the bias, variance, and MSE of estimating the correct coefficient of 1 in Tab. 1(b).

Table 2: Results of the convolutional example

|  | (a) ATT | | | (b) CATT | | |
| --- | --- | --- | --- | --- | --- | --- |
| Method | Bias | SE | RMSE | Bias | SE | RMSE |
| Raw | 3.90 | 0.06 | 3.90 | -0.44 | 0.03 | 0.44 |
| IPW | 6.43 | 0.06 | 6.43 | 1.05 | 0.01 | 1.05 |
| IPWn | 3.32 | 0.41 | 3.35 | -0.39 | 0.14 | 0.42 |
| Regn | 2.61 | 0.17 | 2.62 | -0.33 | 0.06 | 0.34 |
| AIPW | 2.61 | 0.17 | 2.62 | – | – | – |
| AIPWn | 2.61 | 0.13 | 2.61 | – | – | – |
| $DM_0$ | 2.24 | 0.96 | **2.44** | -0.30 | 0.12 | 0.32 |
| $DM_{\frac{1}{2}}$ | 2.49 | 0.65 | 2.57 | -0.28 | 0.08 | **0.30** |
| $DM_0$ DR | 2.59 | 0.14 | 2.59 | – | – | – |
| $DM_{\frac{1}{2}}$ DR | 2.59 | 0.14 | 2.60 | – | – | – |

## 5.2 A Convolutional Example

We next consider an example with confounding image data and solutions based on CNNs. For this, we use the MNIST dataset (LeCun, 1998). To generate the covariate data, we draw $n$ digits uniformly at random. For each one we uniformly draw a random image labeled with that digit from the MNIST dataset and let the pixels $X_i \in \mathbb{R}^{28 \times 28}$ be the covariates. Let $N(X) \in \{0, \ldots, 9\}$ be the digit corresponding to the image and let $B(X)$ be the mean pixel brightness of the image. Then to introduce a complex form of confounding, for each digit 0–4 separately, we take the 10% lightest images (smallest $B(X)$) and label them as treated $T_i = 1$ and the rest as untreated $T_i = 0$. We do the same for 5–9 but take 90% lightest images as treated. We generate outcomes as $Y_i(t) = \text{clip}_{[0,9]}(N(X) + \epsilon_i) + T_i B(X) - T_i$ where $\epsilon_i \sim \text{Uniform}\{-1, 0, 1\}$. We consider all the same estimation strategies as in the last example but use CNNs for all neural networks. Specifically, we use two $5 \times 5$-convolution layers with volume depths 32 and 64 followed by a fully connected layer of 1024 neurons. We run 50 replications of the experiment with $n = 1000$ and report errors in estimating either ATT or CATT in $X^H = B(X)$ in Tab. 2. The results demonstrate that DeepMatch improves causal estimates in settings with rich data and complex confounding.

# 6  Conclusions

Training balancing weights against an adversarial discriminator, each represented by a neural network, provided a way to balance covariates that may be rich and have complex relationships with outcomes and treatments – a challenging setting to standard approaches. Theory and empirics demonstrated that DeepMatch yields stable balancing weights even when an appropriate representation is unavailable a priori.

# 7  Proofs

*Proof of Thm. 1.* Note that $t = 0$ is feasible in eq. (3.1) and, since $\ell(0) = \log(1/2) + \log(2) = 0$, it gives objective value 0, so $\mathrm{DD}^2_{\mathcal{F},\psi}(S^+, S^-) \geq 0$. On the other hand, $\ell(\cdot) \leq \log(2)$ so that the objective in eq. (3.1) is bounded above: $\mathrm{DD}^2_{\mathcal{F},\psi}(S^+, S^-) \leq \sum_{s=\pm} \sum_{i=1}^{n^s} w_i^s \log(2) < \infty$. □

*Proof of Thm. 2.* Letting

$$\mathcal{F}_{|\mathbf{x}} = \{(f(x_i^s))_{s=\pm, i \leq n^s}) : f \in \mathcal{F} \cup -\mathcal{F}\}, \quad \overline{\mathcal{F}}_{|\mathbf{x}} = \{(t, t'\mathbf{f}) : f \in \mathcal{F}_{|\mathbf{x}}, t \geq t' \geq 0\},$$

we can rewrite $\mathrm{DD}^2_{\mathcal{F},\psi}(\{(w_i^+, x_i^+)\}_{i=1}^{n^+}, \{(w_i^-, x_i^-)\}_{i=1}^{n^-})$ as

$$\sup_{(t,\mathbf{f}) \in \overline{\mathcal{F}}_{|\mathbf{x}}, \mathbf{z} \in \mathbb{R}^{n^+ + n^-} : z_i^s = s f_i^s} \sum_{s=\pm} \sum_{i=1}^{n_+} w_i^s \ell(z_i^s) - \frac{\psi}{2} t^2. \quad (7.1)$$

First, we show that $\overline{\mathcal{F}}_{|\mathbf{x}} \subset \mathbb{R}^{n^+ + n^-}$ is convex so that eq. (7.1) is a convex program. Let $(t_0, (t'_0 \mathbf{f}_0)), (t_1, (t'_1 \mathbf{f}_1)) \in \overline{\mathcal{F}}_{|\mathbf{x}}$. Letting $t = \lambda t_0 + (1 - \lambda)t_1$, $t' = \lambda t'_0 + (1 - \lambda)t'_1$, and $\mathbf{f} = (\lambda t'_0/t')\mathbf{f}_0 + (1 - \lambda t'_0/t')\mathbf{f}_1$, we have $t \geq t'$, $\mathbf{f} \in \mathcal{F}_{|\mathbf{x}}$ due to convexity of $\mathcal{F}$, and $t\mathbf{f} = \lambda t_0 \mathbf{f}_0 + (1 - \lambda)t_1 \mathbf{f}_1$. Next we show that $(1, \mathbf{0}) \in \mathrm{relint}(\overline{\mathcal{F}}_{|\mathbf{x}})$. Let any $(t, t'\mathbf{f}) \in \overline{\mathcal{F}}_{|\mathbf{x}}$ be given. Let $\mu = 2t/(2t - 1)$ if $t > 1$ and otherwise $\mu = 2$. Then $\mu \in (1, 2]$ and $\mu/(2\mu - 1) \geq t$ so that $(1 - \mu)t + \mu \geq (\mu - 1)t \geq (\mu - 1)t' \geq 0$. Note that $-\mathbf{f} \in \mathcal{F}_{|\mathbf{x}}$ by symmetry of $\mathcal{F}$. Therefore,

$$(1 - \mu)(t, t'\mathbf{f}) + \mu(1, \mathbf{0}) = ((1 - \mu)t + \mu, (\mu - 1)t'(-\mathbf{f})) \in \overline{\mathcal{F}}_{|\mathbf{x}}$$

So, by Thm. 6.4 of Rockafellar (1997), $(1, \mathbf{0}) \in \mathrm{relint}(\overline{\mathcal{F}}_{|\mathbf{x}})$. Letting $\mathbf{z}$ be defined by its constraint, we then have a Slater point. Note that $\ell_*(p) =$

$\inf_z(pz - \ell(z)) = -h(p) - \log(2)$ for $p \in [0, 1]$ and otherwise $\ell_*(p) = -\infty$. By Prop. 5.3.2 (strong duality) of Bertsekas (1999), we have that eq. (7.1) is equal to

$$\inf_{\gamma \in \mathbb{R}^{n_+ + n_-}} \sup_{(t,\mathbf{f}) \in \overline{\mathcal{F}}_{|\mathbf{x}}, \mathbf{z} \in \mathbb{R}^{n_+ + n_-}} \sum_{s=\pm} \sum_{i=1}^{n_+} (w_i^s \ell(z_i^s) - \gamma_i^s z_i^s) + \sum_{s=\pm} \sum_{i=1}^{n_+} s\gamma_i^s f_i^s - \frac{\psi}{2} t^2$$

$$= \inf_{\mathbf{p} \in \mathbb{R}^{n_+ + n_-}} \sum_{s=\pm} \sum_{i=1}^{n_+} \sup_{z_i^s} w_i^s(\ell(z_i^s) - p_i^s z_i^s)$$

$$+ \sup_{t \geq t' \geq 0} \left( t' \sup_{f \in \mathcal{F}_{|\mathbf{x}}} \sum_{s=\pm} \sum_{i=1}^{n_+} sw_i^s p_i^s f_i^s - \frac{\psi}{2} t^2 \right)$$

$$= \inf_{0 \leq \mathbf{p} \leq 1} \sum_{s=\pm} \sum_{i=1}^{n_+} w_i^s h(p_i^s)$$

$$+ \sup_{t \geq 0} \left( t \operatorname{IPM}_{\mathcal{F}}(\{(p_i^+ w_i^+, x_i^+)\}_{i=1}^{n_+}, \{(p_i^- w_i^-, x_i^-)\}_{i=1}^{n_-}) - \frac{\psi}{2} t^2 \right),$$

which yields the stated result for the two cases, $\psi > 0$ and $\psi = 0$, after taking sup over $t \geq 0$. □

*Proof of Thm. 3.* Fix $\mathbf{x}$ and $\mathbf{w}$. Write $\operatorname{DD}_{\mathcal{F},\psi}$ and $\operatorname{IPM}_{\mathcal{F}}^2$ for shorthand for the distances between the two weighted samples and let

$$H^2(\mathbf{p}) = \sum_{i=1}^{n_+} w_i^+ h(p_i^+) + \sum_{i=1}^{n_-} w_i^- h(p_i^-),$$

$$D^2(\mathbf{p}) = \operatorname{IPM}_{\mathcal{F}}^2(\{(p_i^+ w_i^+, x_i^+)\}_{i=1}^{n_+}, \{(p_i^- w_i^-, x_i^-)\}_{i=1}^{n_-}),$$

so that $\operatorname{DD}_{\mathcal{F},\psi} = \inf_{0 \leq \mathbf{p} \leq 1} H^2(\mathbf{p}) + (\psi^{-1}/2) D^2(\mathbf{p})$. Note that for $p \in [0, 1]$, $D^2(p\mathbf{1}) = p^2 \operatorname{IPM}_{\mathcal{F}}^2$. Note also that $h(p)$ is 4-strongly convex with a unique minimum at $p = 1/2$ so that $h(p) \geq 2(p-1/2)^2 \geq 0$ and $h(p) = 0 \iff p = 1/2$. Hence, the objective in eq. (3.2) is nonnegative for all $\mathbf{p}$ and therefore, by Thm. 2, $\operatorname{DD}_{\mathcal{F},\psi}^2 \geq 0$ and so $\operatorname{DD}_{\mathcal{F},\psi}$ is its well-defined square-root or infinite when $\operatorname{DD}_{\mathcal{F},\psi}^2$ is infinite.

Now suppose $\operatorname{IPM}_{\mathcal{F}} = 0$. Then $D^2(\mathbf{1}/2) = \operatorname{IPM}_{\mathcal{F}}^2/4 = 0$ and $H^2(\mathbf{1}/2) = 0$. Therefore, since $\mathbf{p} = \mathbf{1}/2$ is feasible in eqs. (3.2) and (3.3), we have that $\operatorname{DD}_{\mathcal{F},\psi}^2 = 0$.

Now suppose $\operatorname{DD}_{\mathcal{F},\psi}^2 = 0$. Then there exists $\mathbf{p}$ such that $D^2(\mathbf{p}) = H^2(\mathbf{p}) = 0$ since both functions are nonnegative. Since $H^2(\mathbf{p}) = 0$ and $w_i^s h(p_i^2)$ are nonnegative functions, we have $w_i^s h(p_i^s) = 0$ and hence $w_i^s > 0 \implies p_i^s =$ 1/2 and consequently $w_i^s p_i^s = w_i^s/2$. Therefore, $0 = D^2(\mathbf{p}) = D^2(\mathbf{1}/2) = \text{IPM}_\mathcal{F}^2 /4$. □

*Proof of Thm. 4.* Reusing the notation from the proof of Thm. 3, for any $\mathbf{p}$, we have

$$\left| D(\mathbf{p}) - \frac{1}{2}\text{IPM}_\mathcal{F} \right| = \sup_{f \in \mathcal{F}} \left| \sum_{s=\pm} \sum_{i=1}^{n^s} s w_i^s p_i^s f(x_i^s) \right| - \frac{1}{2} \sup_{f \in \mathcal{F}} \left| \sum_{s=\pm} \sum_{i=1}^{n^s} s w_i^s f(x_i^s) \right|$$

$$\leq \sup_{f \in \mathcal{F}} \left| \sum_{s=\pm} \sum_{i=1}^{n^s} s w_i^s \left( p_i^s - \frac{1}{2} \right) f(x_i^s) \right|$$

$$\leq M \sum_{s=\pm} \sum_{i=1}^{n^s} w_i^s \left| p_i^s - \frac{1}{2} \right|$$

$$\leq M \sqrt{\sum_{s=\pm} \sum_{i=1}^{n^s} w_i^s} \sqrt{\sum_{s=\pm} \sum_{i=1}^{n^s} w_i^s \left( p_i^s - \frac{1}{2} \right)^2}$$

$$\leq M \sqrt{\sum_{s=\pm} \sum_{i=1}^{n^s} w_i^s} \sqrt{\frac{1}{2} \sum_{s=\pm} \sum_{i=1}^{n^s} w_i^s h(p_i^s)}$$

$$\leq \frac{M\sqrt{\overline{w}}}{\sqrt{2}} H(\mathbf{p}).$$

First, consider $\psi = 0$. Let $\mathbf{p}$ be such that $\text{DD}_{\mathcal{F},\psi} = H(\mathbf{p})$ and $D(\mathbf{p}) = 0$. Applying the above bound, we get

$$\text{IPM}_\mathcal{F} \leq \sqrt{2} M \sqrt{\overline{w}} \, \text{DD}_{\mathcal{F},\psi}$$
$$\leq 2 M \sqrt{\overline{w}} \, \text{DD}_{\mathcal{F},\psi}$$
$$\leq \max(2M\sqrt{\overline{w}}, 4\sqrt{\psi}) \, \text{DD}_{\mathcal{F},\psi}.$$

Next, consider $\psi > 0$. Let $\mathbf{p}$ be such that $\text{DD}_{\mathcal{F},\psi}^2 = H(\mathbf{p}) + \psi^{-1} D^2(\mathbf{p})/2$. Applying Jensen's inequality to the above bound, we get

$$\text{IPM}_\mathcal{F} \leq M\sqrt{2\overline{w}} H(\mathbf{p}) + 2D(\mathbf{p})$$
$$\leq \sqrt{4M^2 \overline{w} H^2(\mathbf{p}) + 8D^2(\mathbf{p})}$$
$$= \sqrt{4M^2 \overline{w} H^2(\mathbf{p}) + 16\psi \psi^{-1} D^2(\mathbf{p})/2}$$
$$\leq \max(2M\sqrt{\overline{w}}, 4\sqrt{\psi}) \, \text{DD}_{\mathcal{F},\psi}.$$

On the other hand, because $\mathbf{1}/2$ is feasible in eq. (3.2),

$$8\psi\,\mathrm{DD}_{\mathcal{F},\psi}^2 \leq 8\psi H^2(\mathbf{1}/2) + 4D^2(\mathbf{1}/2) = \mathrm{IPM}_{\mathcal{F}}^2$$

and so $2\sqrt{2\psi}\,\mathrm{DD}_{\mathcal{F},\psi} \leq \mathrm{IPM}_{\mathcal{F}}$. $\square$

*Proof of Thm. 5.* We reuse the notation from the proof of Thm. 3. Thm. 4 showed that $2\sqrt{2\psi}\,\mathrm{DD}_{\mathcal{F},\psi} \leq \mathrm{IPM}_{\mathcal{F}}$. To prove the present result, we further show that whenever $\psi \geq \overline{w}M^2$, we have $\sqrt{1 - \psi^{-1}\overline{w}M^2}\,\mathrm{IPM}_{\mathcal{F}} \leq 2\sqrt{2\psi}\,\mathrm{DD}_{\mathcal{F},\psi}$. Toward that end, let $D^2(\mathbf{p}) = H^2(\mathbf{p}) + \psi^{-1}D^2(\mathbf{p})/2$ and let $\mathbf{p}^*$ be such that $\mathrm{DD}_{\mathcal{F},\psi}^2 = D^2(\mathbf{p}^*)$. Let $W \in \mathbb{R}^{(n^+ + n^-) \times (n^+ + n^-)}$ be the diagonal matrix with $w_i^s$ on its diagonal. WLOG, $W \succ 0$ because otherwise we can consider the equivalent problem after removing all the points with zero weight. Note that $H^2$ is a convex twice-differentiable function, that $\frac{\partial^2}{\partial \gamma^2} h(p) = 1/p + 1/(1-p) \geq 4$, and therefore $\nabla^2 H^2(\mathbf{p})$ is diagonal with entries bounded below by $4w_i^s$. Since $D^2$ is also convex, we have that $D^2(\mathbf{p}) - 2(\mathbf{p} - \mathbf{p}^*)^T W(\mathbf{p} - \mathbf{p}^*)$ is convex and therefore also has an optimum at $\mathbf{p}^*$. Therefore,

$$D^2(\mathbf{p}^*) \leq D^2(\mathbf{1}/2) - 2(\mathbf{p} - \mathbf{1}/2)^T W(\mathbf{p} - \mathbf{1}/2).$$

Let $g \in \partial D^2(\mathbf{1}/2)$ be any subderivative of $D^2$ at $\mathbf{1}/2$. Then, by Cauchy-Schwartz and the above,

$$\begin{aligned}
D^2(\mathbf{1}/2) - D^2(\mathbf{p}^2) &\leq g^T(\mathbf{p}^* - \mathbf{1}/2) \\
&= g^T W^{-1/2} W^{1/2}(\mathbf{p}^* - \mathbf{1}/2) \\
&\leq \sqrt{g^T W^{-1} g}\sqrt{(\mathbf{p}^* - \mathbf{1}/2)W(\mathbf{p}^* - \mathbf{1}/2)} \\
&\leq \sqrt{g^T W^{-1} g}\sqrt{D^2(\mathbf{1}/2) - D^2(\mathbf{p}^*)}/\sqrt{2}.
\end{aligned}$$

Therefore,
$$D^2(\mathbf{1}/2) - D^2(\mathbf{p}^*) \leq g^T W^{-1} g/2.$$

Note that $\partial H^2(\mathbf{1}/2) = 0$ so that $\partial D^2(\mathbf{1}/2) = (2\psi)^{-1} \partial D^2(\mathbf{1}/2) = \psi^{-1} D(\mathbf{1}/2) \partial D(\mathbf{1}/2) = (2\psi)^{-1}\,\mathrm{IPM}_{\mathcal{F}}\,\partial D(\mathbf{1}/2)$. Moreover,

$$\left|(\partial D(\mathbf{p}))_{s,i}\right| = \left|\left(\partial_{\mathbf{p}} \sup_{f \in \mathcal{F}_{|\mathbf{x}}} \left|\sum_{s,i} s w_i^s p_i^s f_i^s\right|\right)_{s,i}\right|$$

$$= \left|\left(\partial_{\mathbf{p}} \sup_{f \in \mathcal{F}_{|\mathbf{x}} \cup -\mathcal{F}_{|\mathbf{x}}} \sum_{s,i} s w_i^s p_i^s f_i^s\right)_{s,i}\right|$$

$$\leq w_i^s \sup_{f \in \mathcal{F}_{|\mathbf{x}} \cup -\mathcal{F}_{|\mathbf{x}}} |f_i^s| \leq w_i^s M.$$

Therefore,

$$\frac{\text{IPM}_\mathcal{F}^2}{8\psi} - \text{DD}_{\mathcal{F},\psi}^2 = D^2(\mathbf{1}/2) - D^2(\mathbf{p}^*) \leq g^T W^{-1} g/2 \leq \frac{\text{IPM}_\mathcal{F}^2}{8\psi^2} \overline{w} M^2,$$

which, when $\psi \geq \overline{w}M^2$, yields $2\sqrt{2\psi} \, \text{DD}_{\mathcal{F},\psi}^2 \geq \sqrt{1 - \psi^{-1}\overline{w}M^2} \, \text{IPM}_\mathcal{F}$ and hence the result. □

*Proof of Thm. 6.* Since $\mathcal{W}$ is a convex set, it remains only to be shown that the objective is convex. The second term, $\frac{\lambda}{n_1}\|W\|_2^2$, is clearly convex when $\lambda \geq 0$. The first term is the supremum over linear forms in $W$ and is therefore also convex. □

*Proof of Thm. 7.* By Thm. 6, eq. (3.4) is convex. Since $W_i = n_1/n_0$ gives a Slater point, strong duality holds and there exists an optimal Lagrange multiplier $\phi^*$ such that eq. (3.4) is equivalent to $\mathcal{W}(\phi^*)$ and any optimizer $W^* \in \mathcal{W}(\phi^*)$ also optimizes eq. (3.4). It must therefore already satisfy the constraint $\sum_{i \in \mathcal{T}_0} W_i^* = n_1$ and hence its objective in eq. (4.1) is exactly the same as in eq. (3.4). For any other $W \in \bigcup_{\phi \geq 0} \mathcal{W}(\phi)$, the objective in eq. (4.1) is the same as the objective of $\frac{n_1 W}{\sum_{i \in \mathcal{T}_0} W_i}$ in eq. (3.4), where the latter is feasible. So. by its optimality in eq. (3.4), $W^*$ also optimizes eq. (4.1). □

We use the following lemma in the proof of Thm. 8.

**Lemma 9.** *For nonnegative random variables $Z_n \geq 0$ and any sub-sigma algebra $\mathcal{G}$,*
$$\mathbb{E}[Z_n \mid \mathcal{G}] = O_p(1) \implies Z_n = O_p(1).$$

Note that when $\mathcal{G} = \sigma(Z_1, \dots)$ is the complete sigma algebra then $\mathbb{E}[Z_n \mid \mathcal{G}] = Z_n$ and the result is trivial and when $\mathcal{G} = \{\varnothing, \Omega\}$ is the trivial sigma algebra then $\mathbb{E}[Z_n \mid \mathcal{G}] = \mathbb{E}[Z_n]$ and the result is direct from Markov's inequality. The lemma provides a proof for the in-between cases.

*Proof of Lemma 9.* Suppose $\mathbb{E}[Z_n \mid \mathcal{G}] = O_p(1)$. Let $\nu > 0$ be given. Then $\mathbb{E}[Z_n \mid \mathcal{G}] = O_p(1)$ says that there exist $N, M$ such that $\mathbb{P}(\mathbb{E}[Z_n \mid \mathcal{G}] > M) \leq \nu/2$ for all $n \geq N$. Let $M_0 = \max\{M, 2/\nu\}$. Then, for all $n \geq N$,

$$\begin{aligned}
\mathbb{P}(Z_n > M_0^2) &= \mathbb{P}(Z_n > M_0^2, \mathbb{E}[Z_n \mid \mathcal{G}] > M_0) + \mathbb{P}(Z_n > M_0^2, \mathbb{E}[Z_n \mid \mathcal{G}] \leq M_0) \\
&= \mathbb{P}(Z_n > M_0^2, \mathbb{E}[Z_n \mid \mathcal{G}] > M_0) + \mathbb{E}[\mathbb{P}(Z_n > M_0^2 \mid \mathcal{G})\mathbb{I}[\mathbb{E}[Z_n \mid \mathcal{G}] \leq M_0]] \\
&\leq \nu/2 + \mathbb{E}\left[\frac{\mathbb{E}[Z_n \mid \mathcal{G}]}{M_0^2}\mathbb{I}[\mathbb{E}[Z_n \mid \mathcal{G}] \leq M_0]\right] \leq \nu/2 + 1/M_0 \leq \nu.
\end{aligned}$$
□

*Proof of Thm. 8.* Let $Z = \sum_{i \in \mathcal{T}_0} \frac{e(X_i)}{1-e(X_i)}$ and $\tilde{W}_i = \frac{n_1}{Z} \frac{e(X_i)}{1-e(X_i)}$. First, note that by assumption $\tilde{W} \in \mathcal{W}(\Theta_g)$ is feasible. Moreover,

$$\text{IPM}_{\mathcal{F}}(\{(\frac{1}{n_1}, X_i)\}_{i \in \mathcal{T}_1}, \{(\frac{\tilde{W}_i}{n_1}, X_i)\}_{i \in \mathcal{T}_0})$$

$$= \sup_{f \in \mathcal{F}} \left| \sum_{i=1}^{n} \left( \frac{T_i}{n_1} - \frac{(1-T_i)e(X_i)}{Z(1-e(X_i))} \right) f(X_i) \right|$$

$$= \frac{n_1}{Z} \sup_{f \in \mathcal{F}} \left| \frac{1}{n_1} \sum_{i=1}^{n} \left( \frac{ZT_i}{n_1} - \frac{(1-T_i)e(X_i)}{(1-e(X_i))} \right) f(X_i) \right|$$

$$\leq \frac{n_1}{Z} \sup_{f \in \mathcal{F}} \left| \frac{1}{n_1} \sum_{i=1}^{n} \left( \frac{ZT_i}{n_1} - T_i \right) f(X_i) \right|$$

$$+ \frac{n_1}{Z} \sup_{f \in \mathcal{F}} \left| \frac{1}{n_1} \sum_{i=1}^{n} \left( T_i - \frac{(1-T_i)e(X_i)}{(1-e(X_i))} \right) f(X_i) \right|$$

$$\leq \underbrace{\frac{n_1}{Z} M \left| \frac{Z}{n_1} - 1 \right|}_{A} + \underbrace{\frac{n_1}{Z} \frac{n}{n_1} \sup_{f \in \mathcal{F}} \left| \frac{1}{n} \sum_{i=1}^{n} \left( T_i - \frac{(1-T_i)e(X_i)}{(1-e(X_i))} \right) f(X_i) \right|}_{B}$$

We will next bound each of the two terms, $A$ and $B$.

Note that $\mathbb{E}[Z] = n_1$. By assumption, there exists $\eta$ with $1 - e(x) > \eta > 0$. Therefore, since $\mathbb{E}[e^2(X_i)/(1-e(X_i))^2] < 1/\eta^2$, Chebychev's inequality yields $|Z - n_1| = O_p(\sqrt{n})$ and hence $\left|\frac{Z}{n_1} - 1\right| = O_p(1/\sqrt{n})$. Since this also means that $n_1/Z \to_p 1$, Slutsky's theorem yields that $A = O_p(1/\sqrt{n})$.

Next, let $\Delta_n = \sup_{f \in \mathcal{F}} \left| \frac{1}{n} \sum_{i=1}^{n} \left( T_i - \frac{(1-T_i)e(X_i)}{(1-e(X_i))} \right) f(X_i) \right|$. Note that $\Delta_n$ satisfies bounded differences with $c_i = 2M/(n\eta)$. Therefore, by McDiarmid's inequality, $\Delta_n \leq \mathbb{E}\Delta_n + O_p(1/\sqrt{n})$. Since $\mathbb{E}\left[T_i - \frac{(1-T_i)e(X_i)}{(1-e(X_i))}\right] = 0$, symmetrization yields that

$$\mathbb{E}\Delta_n \leq 2\mathbb{E}_{T_{1:n}, X_{1:n}} \mathbb{E}_{\xi_{1:n}} \sup_{f \in \mathcal{F}} \left| \frac{1}{n} \sum_{i=1}^{n} \xi_i \left( T_i - \frac{(1-T_i)e(X_i)}{(1-e(X_i))} \right) f(X_i) \right|,$$

where $\xi_i$ are iid Rademacher random variables. Finally, by the Ledoux-Talagrand comparison lemma (Ledoux & Talagrand, 1991, Thm. 4.12), since $\left|T_i - \frac{(1-T_i)e(X_i)}{(1-e(X_i))}\right| \leq 1/\eta$ we have from the above that

$$\mathbb{E}\Delta_n \leq \frac{2}{\eta} \mathbb{E}\mathfrak{R}_n(\mathcal{F}),$$

Finally, since $\mathfrak{R}_n(\mathcal{F})$ satisfies bounded differences itself with $c_i = 2M/n$, McDiarmid's inequality yields that $\mathbb{E}\mathfrak{R}_n(\mathcal{F}) \leq \mathfrak{R}_n(\mathcal{F}) + O_p(1/\sqrt{n})$. Since $n_1/Z \to_p 1$ and $n/n_1 \to_p 1/\mathbb{P}(T_1 = 1)$, Slutsky's theorem yields $B = O_p(\mathfrak{R}_n(\mathcal{F}) + 1/\sqrt{n})$.

All together, we see that $\mathrm{IPM}_\mathcal{F}(\{(\frac{1}{n_1}, X_i)\}_{i \in \mathcal{T}_1}, \{(\frac{\tilde{W}_i}{n_1}, X_i)\}_{i \in \mathcal{T}_0}) = O_p(\mathfrak{R}_n(\mathcal{F}) + 1/\sqrt{n})$. Consequently, for any $\psi > 0$, by Thm. 4, we have that

$$\mathrm{DD}_{\mathcal{F},\psi}(\{(\tfrac{1}{n_1}, X_i)\}_{i \in \mathcal{T}_1}, \{(\tfrac{\tilde{W}_i}{n_1}, X_i)\}_{i \in \mathcal{T}_0}) = O_p(\mathfrak{R}_n(\mathcal{F}) + 1/\sqrt{n}).$$

Next, note that since $\mathbb{E}\left[e^4(X_i)/(1-e(X_i))^4\right] < 1/\eta^4$, Chebychev's inequality yields $\frac{1}{n}\sum_{i=1}^n e^2(X_i)/(1-e(X_i))^2 = O_p(1)$, which, since $n_1^2/Z^2 \to_p 1$, yields by Slutsky's theorem that $\|\tilde{W}\|_2^2 = O_p(n)$. We conclude that the objective of $\tilde{W}$ in eq. (3.4) satisfies

$$\mathrm{DD}_{\mathcal{F},\psi}^2(\{(\tfrac{1}{n_1}, X_i)\}_{i \in \mathcal{T}_1}, \{(\tfrac{\tilde{W}_i}{n_1}, X_i)\}_{i \in \mathcal{T}_0}) + \tfrac{\lambda}{n_1^2}\|\tilde{W}\|_2^2 = O_p(\mathfrak{R}_n^2(\mathcal{F}) + 1/n).$$

Since $W^*$ is optimal and $\tilde{W}$ is feasible, we must therefore also have

$$\mathrm{DD}_{\mathcal{F},\psi}^2(\{(\tfrac{1}{n_1}, X_i)\}_{i \in \mathcal{T}_1}, \{(\tfrac{W_i^*}{n_1}, X_i)\}_{i \in \mathcal{T}_0}) + \tfrac{\lambda}{n_1^2}\|W^*\|_2^2 = O_p(\mathfrak{R}_n^2(\mathcal{F}) + 1/n),$$

and, since each term is nonnegative and by Thm. 4 for $\psi > 0$, consequently

$$\mathrm{IPM}_\mathcal{F}(\{(\tfrac{1}{n_1}, X_i)\}_{i \in \mathcal{T}_1}, \{(\tfrac{W_i^*}{n_1}, X_i)\}_{i \in \mathcal{T}_0}) = O_p(\mathfrak{R}_n(\mathcal{F}) + 1/\sqrt{n}),$$
$$\tfrac{1}{n_1^2}\|W^*\|_2^2 = O_p(\mathfrak{R}_n^2(\mathcal{F}) + 1/n).$$

Because $f_0 = f(x; \theta_f)$ and $R(\cdot)$ is degree-2 homogeneous, we have that $|B(W^*, f_0)| \leq R(\theta_f)\mathrm{IPM}_\mathcal{F}(\{(\frac{1}{n_1}, X_i)\}_{i \in \mathcal{T}_1}, \{(\frac{W_i^*}{n_1}, X_i)\}_{i \in \mathcal{T}_0})$ and since $\sigma^2(X) \leq \overline{\sigma}^2$ is almost surely bounded we have $V^2(W; \sigma^2) \leq \overline{\sigma}\left(\frac{1}{n_1} + \frac{1}{n_1^2}\|W^*\|\right)$. By eq. (2.5), $\mathbb{E}\left[(\hat{\tau}_W - \tau_{n_1})^2 \mid X_{1:n}, T_{1:n}\right] = O_p(\mathfrak{R}_n^2(\mathcal{F}) + 1/n)$. Hence, by Jensen's inequality $\mathbb{E}\left[|\hat{\tau}_W - \tau_{n_1}| \mid X_{1:n}, T_{1:n}\right] = O_p(\mathfrak{R}_n(\mathcal{F}) + 1/\sqrt{n})$. Therefore, by Lemma 9, we then must also have that $\hat{\tau}_{W^*} - \tau_{n_1} = O_p(\mathfrak{R}_n(\mathcal{F}) + 1/\sqrt{n})$. Noting that, by LLN, $\tau_{n_1} - \tau = O_p(1/\sqrt{n})$ completes the proof. □